\documentclass{article}

\usepackage{arxiv}

\usepackage[utf8]{inputenc} 
\usepackage[T1]{fontenc}    
\usepackage{hyperref}       
\usepackage{url}            
\usepackage{booktabs}       
\usepackage{amsfonts}       
\usepackage{nicefrac}       
\usepackage{microtype}      
\usepackage{cleveref}       
\usepackage{lipsum}         
\usepackage{graphicx}
\usepackage{doi}

\title{Fair Foundation Models for Medical Image Analysis: Challenges and
Perspectives}

\newif\ifuniqueAffiliation
\uniqueAffiliationfalse

\ifuniqueAffiliation 
\author{ Dilermando Queiroz\thanks{Use footnote for providing further
		information about author (webpage, alternative
		address)---\emph{not} for acknowledging funding agencies.} \\
	Federal University of S\~ao Paulo\\
	Brazil \\
	\texttt{dilermando.queiroz@unifesp.br} \\
	\And
	Anderson Carlos \\
	Federal Institute of Goi\'as \\
	Brazil \\
	\AND
	Andr\'e Anjos \\
	Idiap Research Institute \\
	Switzerland \\
	\And
	Lilian Berton \\
	Federal University of S\~ao Paulo \\
	Brazil \\
}
\else
\usepackage{authblk}

\author[1]{%
	Dilermando Queiroz\thanks{\texttt{dilermando.queiroz@unifesp.br}}%
}
\author[2]{%
	Anderson Carlos%
}
\author[3]{%
	Andr\'e Anjos%
}
\author[1]{%
	Lilian Berton%
}
\affil[1]{Federal University of S\~ao Paulo, Brazil}
\affil[2]{Federal Institute of Goi\'as, Brazil}
\affil[3]{Idiap Research Institute, Switzerland}
\fi

\renewcommand{\shorttitle}{}

\hypersetup{
pdftitle={Fair Foundation Models for Medical Image Analysis: Challenges and
Perspectives},
pdfsubject={q-bio.NC, q-bio.QM},
pdfauthor={Dilermando Queiroz, Anderson Carlos, Andr\'e Anjos, Lilian Berton},
pdfkeywords={Foundation Models, Fairness, World Health},
}

\begin{document}
\maketitle

\begin{abstract}
Ensuring equitable Artificial Intelligence (AI) in healthcare demands systems that make unbiased decisions across all demographic groups, bridging technical innovation with ethical principles.
Foundation Models (FMs), trained on vast datasets through self-supervised learning, enable efficient adaptation across medical imaging tasks while reducing dependency on labeled data.
These models demonstrate potential for enhancing fairness, though significant challenges remain in achieving consistent performance across demographic groups. Our review indicates that effective bias mitigation in FMs requires systematic interventions throughout all stages of development. While previous approaches focused primarily on model-level bias mitigation, our analysis reveals that fairness in FMs requires integrated interventions throughout the development pipeline, from data documentation to deployment protocols. This comprehensive framework advances current knowledge by demonstrating how systematic bias mitigation, combined with policy engagement, can effectively address both technical and institutional barriers to equitable AI in healthcare. The development of equitable FMs represents a critical step toward democratizing advanced healthcare technologies, particularly for underserved populations and regions with limited medical infrastructure and computational resources.
\end{abstract}


\section{Introduction}

\begin{figure*}[h]
    \centering
    \includegraphics[width=1\textwidth]{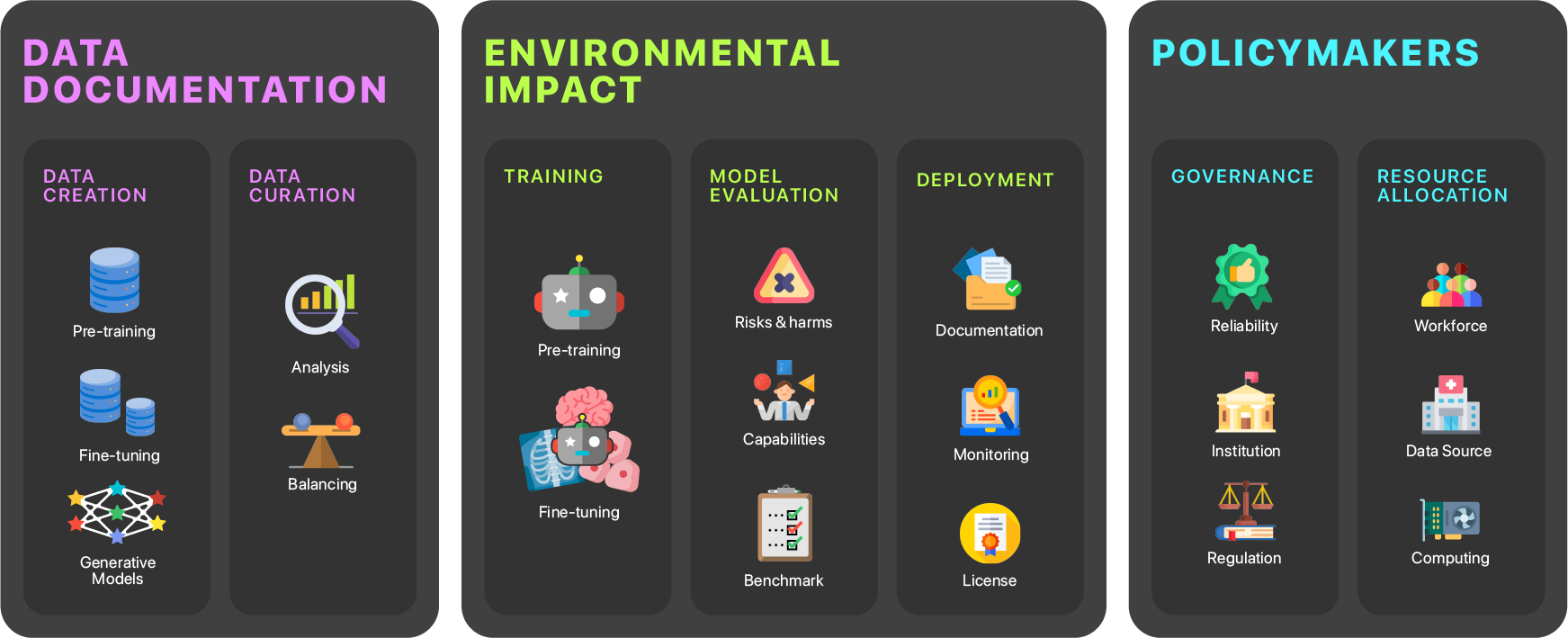}
    \caption{\textbf{Conceptual framework}. This figure delineates sequential phases in FMs development, illustrating principal challenges for achieving fairness. Three focal domains are emphasized: data documentation, curation dataset ensuring diversity and quality to detect and mitigate bias; environmental impact, spanning training, deployment, and resource efficiency; and policymakers, establish governance, standards, and resource distribution to ensure ethical, equitable FMs access.}
    \label{fig:overview}
\end{figure*}

AI in medicine offers transformative potential through improving access to diagnostics and enhancing the speed and quality of medical care \cite{moor_foundation_2023, zhang_challenges_2023}. AI tools extend healthcare services, particularly in resource-limited regions, thereby making care more efficient and accessible \cite{widner_lessons_2023}. However, these advancements raise critical ethical concerns that emphasize the need for a fair and equitable distribution of benefits across all populations \cite{rajkomar_ensuring_2018, riccilara_addressing_2022, mccradden_ethical_2020}. We must develop and apply these technologies responsibly to uphold bioethical principles of justice, autonomy, beneficence, and non-maleficence to prevent discrimination and promote inclusive healthcare for all \cite{beauchamp_principles_1994}.

Governments worldwide are establishing regulatory frameworks for AI across critical sectors. The EU AI Act \cite{europeanparliamentandcouncil_regulation_2024}, the first comprehensive regulation of its kind, has introduced a risk-based classification system for AI applications. In parallel, the U.S. Office for Civil Rights has enacted specific protections against algorithmic bias in healthcare through the Affordable Care Act \cite{unitedstates:nationalarchivesandrecordsadministration:officeofthefederalregister_act_2010}. These regulatory initiatives reflect a coordinated global effort to implement evidence-based guidelines that ensure the fairness, safety, and equity of AI systems \cite{high-levelexpertgrouponartificialintelligence_ethics_2019, longpre_responsible_2024, obermeyer_algorithmic_, lekadir_futureai_2025, vasey_reporting_2022, collins_tripod+ai_2024, tejani_checklist_2024, liu_reporting_2020}.

FMs serve as essential components in AI by enabling diverse tasks across text, image, video, audio, and other domains through their versatility and scalability \cite{bommasani_opportunities_2022}. Through training on massive datasets, these models capture broad patterns within and across domains. However, significant challenges persist in developing ethical models that can effectively identify and reduce inherent biases \cite{khan_how_2023, glocker_risk_2023, li_empirical_2024}. While bias persists across AI systems, FMs demonstrate significant potential for their mitigation \cite{vaidya_demographic_2024, queiroz_using_2025}, thus creating opportunities for unified models that equitably serve diverse populations while driving greater inclusion in AI.

The production of FMs requires substantial resources, including specialized labor, large datasets, and significant computational infrastructure. These high costs restrict development capabilities to select countries, institutions, and companies, thereby increasing the risk of global inequalities. Stakeholders in regions such as Africa have indicated their inability to develop such models, which potentially widens the disparity between populations that benefit from AI and those that do not \cite{ade-ibijola_artificial_2023}. Addressing this challenge requires developing strategies that facilitate the creation of globally accessible FMs while enabling broad participation in their production.

Bias mitigation must occur throughout the proposed development process of FMs, as illustrated in Figure \ref{fig:overview}. This process requires thorough data documentation during the curation and creation phases to ensure dataset diversity and representativeness. The consideration of environmental impacts during training, model evaluation, and deployment phases enhances fairness. Policymakers fulfill a crucial role through the enactment of laws and the allocation of resources that support equitable AI practices. Through the integration of bias mitigation strategies at each developmental stage, researchers can develop inclusive and responsible AI models that serve diverse populations effectively.

A substantial body of research has empirically demonstrated inherent biases in FMs and established initial frameworks for fairness evaluation \cite{jin_fairmedfm_2024, glocker_risk_2023, khan_how_2023, li_empirical_2024}. The FairMedFM benchmark, for instance, provides a standardized basis for these empirical assessments \cite{jin_fairmedfm_2024}. While comprehensive investigations of trustworthiness in medical imaging FMs have been conducted \cite{shi_survey_2024}, these studies primarily address fairness as one component within broader trustworthiness considerations. Additionally, existing analyses provide thorough technical perspectives on medical imaging FMs \cite{zhang_challenges_2023, li_progress_2024} but lack systematic examination of fairness considerations. Current reviews of AI fairness in healthcare \cite{chen_algorithmic_2023, du_fairness_2021, mehrabi_survey_2021, riccilara_addressing_2022, xu_addressing_2024} have not adequately addressed the distinct challenges posed by FMs. To our knowledge, no study has adopted a broader conceptual lens to address the interconnected ethical, environmental, and governance challenges in ensuring fairness in FMs. To address this gap, this review investigates the development lifecycle from a narrative perspective, identifying key opportunities and challenges for fair FMs.

Our contributions are:
\begin{itemize}
\item An investigation of potentials and gaps in bias mitigation methods across all stages of FMs development.
\item A global assessment of dataset distribution patterns reveals inequalities in representation across different regions.
\item An analysis of potential policy interventions for ensuring and promoting the development of more equitable FMs.
\end{itemize}
\section{Background and Taxonomy}
This section delineates the fundamental principles underlying fairness and FMs. We also present a table summarizing key concepts and strategies relevant to our review (Table~\ref{tab:fairness_taxonomy}).

\begin{table*}[t]
\small
\setlength{\tabcolsep}{5pt}
\renewcommand{\arraystretch}{1.5}
\caption{\textbf{Taxonomy.} Key concepts and strategies encompassing fairness, protected attributes, foundation models, vision-language models, and learning approaches relevant to scalable and equitable AI systems.}
\label{tab:fairness_taxonomy}
\centering
\resizebox{\textwidth}{!}{%
\begin{tabular}{p{3.5cm} p{7.5cm} p{6.5cm}}
\toprule
\textbf{Term} & \textbf{Summary} & \textbf{Key Concepts} \\
\midrule
Fairness & The principle that ensures equitable treatment and non-discriminatory outcomes across diverse groups defined by protected attributes in algorithmic decision-making. & Equity, bias mitigation, performance parity, protected groups. \\
Hallucination & The generation of plausible but false or fabricated information not grounded in real-world facts. & Misinformation, clinical risk, reliability, uncertainty, inaccurate knowledge. \\
Protected Attributes & Patient demographic and clinical metadata that require safeguarding to prevent bias in algorithmic decision-making, ensuring fairness across groups defined by these attributes (e.g., age, gender). & Protected attributes, bias prevention, demographic metadata, group fairness. \\
Foundation Models (FMs)& Large-scale pretrained models adaptable to diverse tasks, with emphasis on self-supervised and weakly-supervised training strategies. & Pre-training, downstream task, scalability, robustness. \\
Vision Language Models (VLMs) & Models that integrate visual and linguistic data processing, typically trained using weakly supervised learning to leverage large-scale paired image-text data. & Multimodal learning, weakly supervised learning, visual-linguistic integration, FMs. \\
World Models & Models that simulate and predict the dynamics of the real world, enabling anticipation of future states and informed decision-making. & Environment Simulation, latent space prediction, planning. \\
Generative Models & Models that learn data patterns to create new, original content resembling the training data. & Content generation, hallucinations, synthetic data.\\
Knowledge Agglomeration & Knowledge agglomeration synthesizes knowledge by distilling diverse representations from multiple FMs into a single, unified agglomerative model.
& Knowledge distillation, Model merging, agglomerative models.\\
Self-supervised Learning & Training method using large unlabeled data by leveraging inherent data structures. & Unlabeled data, Masked Autoencoder (MAE) \cite{he_masked_2021}, Simple Framework for Contrastive Learning of Visual Representations (SimCLR) \cite{chen_simple_2020}, Image-based Joint-Embedding Predictive Architecture (I-JEPA) \cite{assran_selfsupervised_2023}. \\
Weakly-supervised Learning & Learning approach that leverages limited, noisy, or imprecise labels to guide model training, commonly used in large-scale and multimodal datasets. & Noisy labels, limited supervision, semi-supervised methods, VLMs, Contrastive Language-Image Pretraining (CLIP)~\cite{radford_learning_2021}, Sigmoid Loss for Language Image Pre-Training (SigLIP)~\cite{zhai_sigmoid_2023}. \\
\bottomrule
\end{tabular}%
}
\end{table*}

\subsection{Fairness}
\textbf{Principles of Trustworthy AI.} The development of medical image analysis AI systems, to achieve trustworthiness, is guided by six fundamental principles: fairness, universality, traceability, usability, robustness, and explainability \cite{lekadir_futureai_2025}. Among these, fairness, which ensures non-discriminatory outcomes across diverse patient populations, represents a critical determinant of ethical AI implementation. Our analysis centers on fairness and examines its interdependencies with other principles, specifically how it interacts with robustness to ensure reliable system performance and with traceability to maintain systematic accountability, thus demonstrating how these principles collectively contribute to trustworthy AI in clinical settings. We specifically center hallucinations as a critical reliability issue, as these generated false but plausible outputs can undermine fairness by causing biased or unequal clinical outcomes.

\textbf{Healthcare Disparities.} Differences in national healthcare delivery capacities result in variable health outcomes across populations \cite{worldhealthorganization_conceptual_2010, chen_ethical_2021}. Within countries that maintain public healthcare systems, disparities persist because of multiple factors, including race, gender, age, ethnicity, body mass index, education, insurance status, and geographic location \cite{bailey_structural_2017, williams_understanding_2019}. These determinants influence an individual's capacity to access treatment, obtain quality care, and achieve favorable health outcomes.

\textbf{AI Bias Manifestation.} AI models frequently assimilate and reproduce biases inherent in their training data, subsequently reflecting and intensifying societal inequalities. These models often depend on spurious correlations, employing computational shortcuts that amplify existing biases \cite{geirhos_shortcut_2020, zou_implications_2023, glocker_algorithmic_2023}. A well-designed fair algorithm produces impartial decisions by ensuring equitable outcomes across demographic groups without discrimination based on protected attributes such as race, gender, or age. Despite the implementation of controlled datasets and balanced groups, bias can manifest through various factors, including image complexity and labeling inconsistencies \cite{cui_classes_2024}. Consequently, unintended biases may emerge even under optimal conditions, highlighting the necessity for continuous evaluation and mitigation procedures to maintain algorithmic fairness.

\textbf{Bias Mitigation Strategies.} Previous research has established three primary classifications for bias mitigation strategies: pre-processing, in-processing, and post-processing \cite{du_fairness_2021, chen_algorithmic_2023, mehrabi_survey_2021, xu_addressing_2024}. Pre-processing approaches modify datasets through demographic representation balancing, protected feature removal (such as race and gender), or synthetic data augmentation to enhance diversity \cite{burlina_addressing_2021, noseworthy_assessing_2020, calmon_optimized_2017}. In-processing methods alter the training process by integrating fairness constraints into the loss function or implementing adversarial training to prevent bias acquisition \cite{celis_improved_2019, wissel_investigation_2019, zafar_fairness_2017}. Post-processing techniques adjust model predictions after training through methods such as prediction calibration to satisfy fairness criteria \cite{kamishima_fairnessaware_2012, pleiss_fairness_2017, chouldechova_case_2018}.

\textbf{Framework Integration.} Our research extends traditional bias mitigation strategies (pre-processing, in-processing, and post-processing) into a comprehensive framework encompassing all stages of FMs development. Through the integration of bias mitigation efforts across all phases and the inclusion of policymakers in technical discussions, as illustrated in Figure 1, this framework systematically addresses biases throughout the development process. The approach responds to significant concerns regarding the potential of these models to amplify global economic inequalities. The incorporation of policymakers ensures the integration of ethical considerations and regulatory measures into FMs development, thus mitigating potential adverse effects on global economic disparities.

\subsection{Foundation Models}

FMs are trained on extensive datasets that can be efficiently adapted to multiple downstream tasks through fine-tuning, thereby eliminating the necessity of training specialized models from scratch. The Vision Transformer (ViT) \cite{dosovitskiy_image_2021} serves as a leading FMs architecture in computer vision for extracting fundamental image features. The integration of self-supervised learning techniques, such as MAE \cite{he_masked_2021} and contrastive learning methods like SimCLR \cite{chen_simple_2020}, enables these models to learn directly from large volumes of unlabeled data, thus largely mitigating the need for costly manual labeling processes. The I-JEPA introduces a self-supervised learning approach that predicts semantic embeddings directly, offering enhanced efficiency compared to pixel-reconstruction methods \cite{assran_selfsupervised_2023}.

\textbf{Capabilities and Applications.} Through the utilization of massive unlabeled datasets, advanced architectures, and self-supervised learning techniques, FMs acquire comprehensive image representations within their training domains. The fine-tuning process facilitates task-specific adaptation through minimal adjustments and limited labeled data requirements. Several key capabilities establish their essential role in scalable, real-world applications: rapid task adaptation, robustness to distribution shifts, and efficient utilization of labeled data \cite{zhou_foundation_2023, azizi_robust_2023, wu_generalist_2023, tu_generalist_2023}. For example, FMs improve diagnostic accuracy by 11.5\% relative to supervised methods and, in out-of-distribution settings, achieve comparable performance using only 1–33\% of the labeled data \cite{azizi_robust_2023}.

\textbf{Multimodal Integration.} These models facilitate training across diverse data types within a unified architectural framework. VLMs demonstrate this capability by integrating visual and linguistic processing through weakly supervised learning that combines image captions with contrastive losses, such as CLIP \cite{radford_learning_2021}. This integration supports simultaneous interpretation and generation of information across modalities. Subsequent work has improved training by replacing contrastive loss with a sigmoid-based loss, as in SigLIP \cite{zhai_sigmoid_2023a}, while other studies highlight the crucial role of data quality over model architecture or pretraining objectives \cite{xu_demystifying_2023}.

\textbf{Domain Specialization.} Domain-specific FMs constitute specialized architectures tailored to address the distinctive characteristics of particular domains, ranging from natural image processing \cite{oquab_dinov2_2024}, medical imaging applications \cite{wu_generalist_2023, azizi_robust_2023} to VLMs tailored for medical domain \cite{wang_medclip_2022,sellergren_medgemma_2025}. In contrast to general-purpose models, these specialized frameworks concentrate on the unique data structures and analytical challenges inherent to their respective domains. Medical imaging presents a notable example, where image characteristics vary substantially in resolution, ranging from whole-organ visualization to cellular-level structures, with data originating from diverse acquisition modalities including X-ray radiography, computed tomography (CT), magnetic resonance imaging (MRI), and ultrasonography \cite{zhang_challenges_2023}.
\section{Data Documentation}

This section delineates the essential components of data documentation, encompassing both the data creation phase and subsequent curation processes.

\begin{figure*}[h]
    \centering
    \includegraphics[width=1\textwidth]{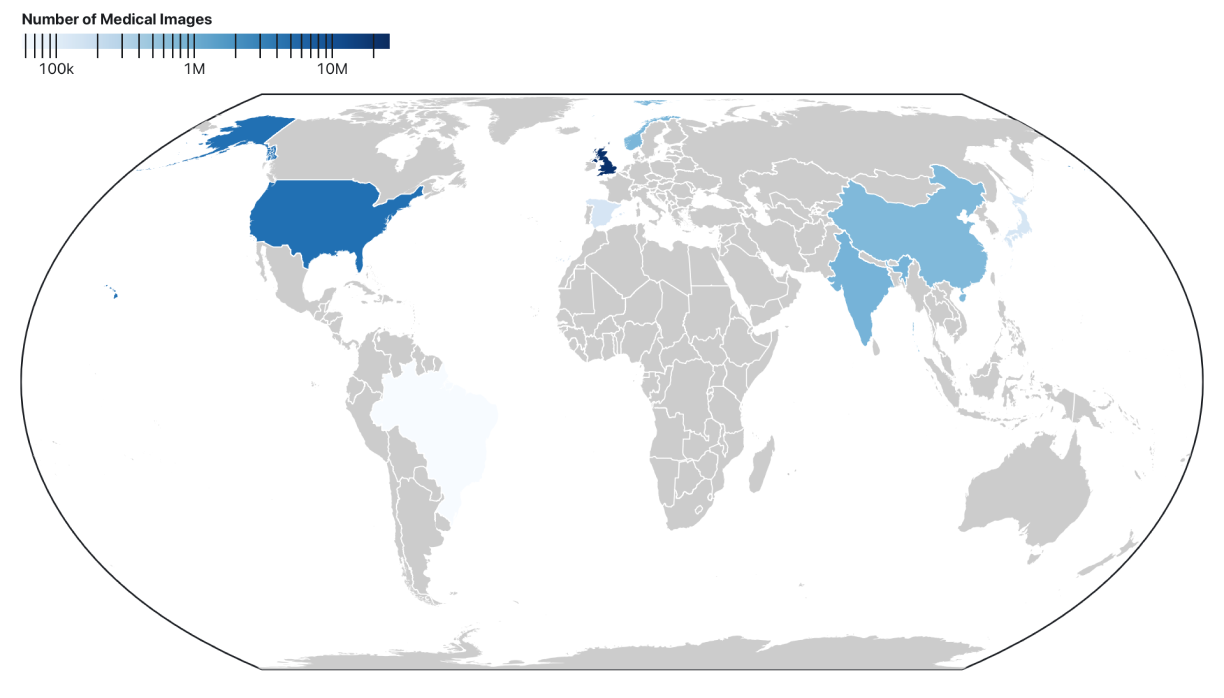}
    \caption{\textbf{Global distribution of medical imaging data:} This geographic visualization depicts the volume of medical imaging datasets by country, excluding multi-country datasets listed in Table \ref{tab:medical-datasets}. The figure highlights pronounced disparities in data representation, underscoring its critical role in the development of equitable FMs. Notably, a limited number of countries account for the majority of available medical imaging data.}
    \label{fig:map}
\end{figure*}

\begin{table*}[t]
\small
\setlength{\tabcolsep}{5pt}
\caption{\textbf{Medical imaging datasets and their characteristics.} Overview of publicly available medical imaging datasets, ordered by size and annotated with key attributes including computational framework, anatomical region, and data accessibility parameters.}
\label{tab:medical-datasets}
\resizebox{\columnwidth}{!}{
\begin{tabular}{l l l l l r l l l}
\toprule
Dataset & Model & Language & Region & Modality & \multicolumn{1}{c}{Images} & Demographics & Origin & License \\
\midrule
Moorfields BioResource 001 \cite{healthdataresearchinnovationgateway_moorfields_2024} & Vision & -- & Ocular & Multimodal & 26,548,820 & A & UK & Restricted \\
MedTrinity-25M\cite{xie_medtrinity25m_2024} & Multimodal & English & Multiple & Multimodal & 25,000,000 & -- & Multiple & Group \\
PMC-15M\cite{zhang_biomedclip_2025} & Multimodal & English & Multiple & Multimodal & 15,000,000 & -- & Multiple & CC BY-SA \\
BRATS24-MICCAI\cite{verdier_2024_2024} & Vision & -- & Cerebral & MRI & 2,535,132 & -- & USA & CC BY 4.0 \\
PMC-OA\cite{lin_pmcclip_2023} & Multimodal & English & Multiple & Multimodal & 1,600,000 & A & Multiple & OpenRAIL \\
RadImageNet\cite{mei_radimagenet_2022} & Vision & -- & Multiple & Multimodal & 1,350,000 & -- & USA & CC BY 4.0 \\
ISIC\cite{zawacki_siimisic_2020} & Vision & -- & Dermal & Dermoscopy & 1,162,456 & A, S & Multiple & CC BY-NC-SA \\
TCGA\cite{kawai_largescale_2023} & Vision & -- & Dermal & Histology & 1,142,221 & A, R, S & USA & CC BY-NC-SA \\
HyperKvasir\cite{borgli_hyperkvasir_2020} & Vision & -- & Colon & Endoscopy & 1,000,000 & -- & Norway & CC BY 4.0 \\
BRATS-ISBI\cite{karargyris_federated_2023} & Vision & -- & Cerebral & MRI & 987,340 & -- & Multiple & CC BY 4.0 \\
BHX\cite{reis_brain_} & Vision & -- & Cerebral & MRI & 973,908 & -- & India & CC BY 4.0 \\
LDPolypVideo\cite{ma_ldpolypvideo_2021} & Vision & -- & Colon & Endoscopy & 901,666 & -- & China & -- \\
MIMIC-CXR-JPG\cite{johnson_mimiccxrjpg_2019} & Multimodal & English & Pulmonary & Radiograph & 370,955 & A, R, S & USA & PHDL \\
CheXpert\cite{irvin_chexpert_2019} & Multimodal & English & Pulmonary & Radiograph & 222,793 & A, R, S & USA & RUA \\
PadChest\cite{bustos_padchest_2020} & Multimodal & Spanish & Pulmonary & Radiograph & 160,868 & A, S & Spain & RUA \\
SUN-SEG\cite{ji_video_2022} & Vision & -- & Colon & Endoscopy & 158,690 & A, S & Japan & MIT \\
NIH-CXR14\cite{wang_chestxray8_2017} & Vision & -- & Pulmonary & Radiograph & 112,120 & A, S & USA & CC0 \\
Kermany et al.\cite{kermany_identifying_2018} & Vision & -- & Ocular & OCT & 108,312 & -- & USA & CC BY 4.0 \\
EyePACS\cite{gulshan_development_2016} & Vision & -- & Ocular & Fundus & 92,501 & -- & India & CC BY 4.0 \\
BRAX\cite{reis_brax_2022} & Vision & -- & Pulmonary & Radiograph & 40,967 & A, S & Brazil & PHDL \\
MURA\cite{rajpurkar_mura_2018} & Vision & -- & Multiple & Radiograph & 40,561 & -- & USA & CC BY 4.0 \\
ASU-Mayo\cite{tajbakhsh_automated_2016} & Vision & -- & Colon & Endoscopy & 19,400 & -- & USA & -- \\
BRSET\cite{nakayama_brset_2024} & Vision & -- & Ocular & Fundus & 16,266 & A, S, N & Brazil & PHDL \\
Harvard-FairVLMed\cite{luo_fairclip_2024} & Multimodal & English & Ocular & SLO & 10,000 & A, R, S, E & USA & CC BY-NC-SA \\
\bottomrule
\multicolumn{8}{p{16cm}}{\footnotesize Demographics: A, Age; R, Race; S, Sex; N, Nationality; E, Ethnicity; --, None reported} \\
\multicolumn{8}{p{16cm}}{\footnotesize Licenses: CC, Creative Commons (BY, Attribution; NC, NonCommercial; SA, ShareAlike); PHDL, PhysioNet Health Data License; RUA, Research Use Agreement} \\
\end{tabular}
}
\end{table*}

\subsection{Data Creation}
The first step in FMs development involves comprehensive data collection. Healthcare facilities generate medical images through specialized imaging equipment that captures various anatomical structures. The pre-training phase utilizes extensive unlabeled datasets, enabling FMs to acquire general representations and discover underlying patterns without annotation requirements. For subsequent fine-tuning processes, labeled data plays an essential role in enabling models to develop task-specific capabilities through the refinement of learned representations. Both unlabeled and labeled datasets must maintain high quality and diversity standards, as these characteristics fundamentally influence FMs performance and generalization capabilities \cite{deitke_molmo_2024, oquab_dinov2_2024}.

\textbf{Inclusion Criteria.} Datasets were identified for inclusion in Table \ref{tab:medical-datasets} through a structured, multi-step process. The initial step involved identifying the pre-training datasets for the principal FMs listed in a recent survey~\cite{shi_survey_2024}. To enhance geographic representation, a supplementary search was conducted on the Hugging Face and PhysioNet platforms to identify datasets from underrepresented regions. A minimum size threshold of 10,000 images was applied to all datasets to ensure they represent large-scale collections. Finally, metadata for each selected dataset were manually extracted from the corresponding publication and official website to populate the table.

While not exhaustive, this methodological approach enabled the systematic curation of 74,184,415 medical images, representing a comprehensive cross-section of diverse imaging modalities and anatomical regions. The comprehensive review encompassed unimodal imaging and multimodal image-text paired datasets, spanning diverse imaging modalities and anatomical regions. This methodological documentation framework facilitates a systematic assessment of the current medical imaging data ecosystem while elucidating opportunities for enhanced demographic representation in FMs development.The geographic distribution of dataset origins appears in Figure \ref{fig:map}.

\textbf{Pre-training.} The utilization of large unlabeled datasets for FMs training significantly reduces dependence on costly and time-intensive labeled data collection. Advanced learning methodologies, particularly self-supervised learning approaches \cite{chen_simple_2020, he_masked_2021, assran_selfsupervised_2023}, facilitate effective utilization of unlabeled data while minimizing labeling bias. This methodology proves especially valuable in medical contexts, where individual clinician annotation preferences, influenced by patient attributes, may introduce systematic biases \cite{berhane_patients_2015}. Using unlabeled data enables FMs to identify generalizable patterns while simultaneously reducing both cost constraints and annotation-induced biases.

The development of robust datasets for FMs presents significant challenges in ensuring comprehensive representation across populations, imaging modalities, equipment specifications, and disease classifications. The accumulation of large-scale data frequently amplifies existing biases and imbalances that reflect underlying disparities in healthcare access and infrastructure. As illustrated in Figure \ref{fig:map}, the predominant source of available datasets resides in developed nations, which substantially constrains the diversity of documented diseases and patient demographics. This geographic concentration of data resources is particularly evident in regions such as Africa, where large-scale medical imaging datasets remain notably absent (see Section \ref{sec:training} for a detailed discussion of the implications of such underrepresentation). 

\textbf{Fine-tuning.} Smaller, more accurate, and more representative datasets are ideal for specializing FMs for specific tasks. This specialization is achieved through supervised learning on labeled datasets. Due to their data efficiency and generalization capabilities \cite{zhou_foundation_2023, azizi_robust_2023, wu_generalist_2023, tu_generalist_2023}, FMs require less data than training supervised models from scratch. Consequently, fine-tuning FMs offers significant advantages, as it requires fewer computational resources and enables countries and institutions to adapt these models to their specific tasks with greater accessibility.

Precise labels and comprehensive patient information, including gender, age, and race, facilitate the evaluation of model performance and enable systematic bias identification \cite{reis_brax_2022, nakayama_brset_2024, irvin_chexpert_2019, groh_evaluating_2021}. This demographic information proves essential for model evaluation, as it enables more rigorous identification and mitigation of potential biases (discussed in the evaluation section \ref{Model Evaluation}). The availability of labeled data supports dataset balancing across specific attributes, thereby enhancing model fairness and performance. This represents a significant advantage over unlabeled data, which presents greater challenges in achieving these objectives \cite{ashurst_fairness_2023}.

\textbf{Generative Models.} Generative models constitute a driving force behind recent advancements in AI and its applications. These models learn to approximate the probability distribution of output features conditioned on input features, enabling the generation of novel data instances that closely approximate the characteristics of the training data. Advanced techniques, including generative adversarial networks (GANs) \cite{liu_radimagegan_2023}, variational autoencoders (VAEs) \cite{pesteie_adaptive_2019}, diffusion models \cite{pinaya_brain_2022}, have significantly enhanced the capabilities of generative AI systems.

Data augmentation through generative approaches facilitates both pre-training and fine-tuning stages. These models enable dataset balancing across protected attributes through synthetic data generation for underrepresented patient groups and enhance model robustness by generating challenging cases, reducing the fairness gap in chest radiograph classifiers trained on synthetic and real images by 44.6\% \cite{cui_classes_2024}. Empirical studies demonstrate that fine-tuning models with augmented data improve both fairness and robustness, resulting in enhanced performance across diverse populations \cite{ktena_generative_2023}. However, generative models may inadvertently amplify biases present in training data, such as generating synthetic medical images that systematically underrepresent darker skin tones \cite{zhang_fairskin_2024}, which necessitates robust bias mitigation strategies when utilizing synthetic data for augmentation.

\textbf{Hallucinations.} Understood as model-generated content that is not aligned with the real world, constitute a major source of misinformation that undermines the reliability of text and image generation models. They are typically associated with high uncertainty and incomplete or inaccurate knowledge \cite{liu_survey_2024, ji_survey_2023, huang_survey_2025, manakul_selfcheckgpt_2023}, particularly in instances where the model lacks relevant training data and exhibits low confidence \cite{kim_medical_2025, xiao_hallucination_2021, liu_survey_2024}. As illustrated in Figure \ref{fig:map}, large regions of the world are underrepresented in current datasets, which implies that models are more prone to hallucinate in these contexts because their underlying knowledge is partial or missing \cite{chen_crosscare_2024}. Nevertheless, it is important to note that hallucinations also occur in settings where the model has the knowledge and model confidence is high \cite{simhi_trust_2025a}, indicating that they remain a persistent concern even under apparently well-specified conditions.

\subsection{Data Curation}
Although increasing data volume in parallel with neural network size can enhance model performance \cite{goyal_selfsupervised_2021, goyal_vision_2022}, the relationship between data quantity and model improvement is not consistently linear. Recent research demonstrates that systematic data curation plays a critical role in both natural language processing and computer vision tasks \cite{deitke_molmo_2024, oquab_dinov2_2024}. Ensuring dataset fairness requires careful attention to multiple dimensions: diversity, global representativeness, equipment variability, disease representation, gender balance, and age distribution. Furthermore, data deduplication serves as an essential process for eliminating near-duplicate images, thereby reducing redundancy and enhancing dataset diversity and representativeness \cite{oquab_dinov2_2024, lee_deduplicating_2022}. In parallel, removing noisy samples, incomplete entries, and ambiguous abbreviations is crucial for mitigating hallucinations and achieve fair representations \cite{hegselmann_datacentric_2024, kim_medical_2025}.

\textbf{Protected attributes.} The acquisition of metadata and labels in medical imaging presents substantial challenges, primarily due to the resource-intensive nature of manual curation given data volume and complexity. While fairness research frequently presumes the availability of demographic data, such information often remains inaccessible due to legal, ethical, and practical constraints \cite{andrus_demographicreliant_2022}. This limited access to demographic data creates significant obstacles for both data curation processes and the development of unbiased models \cite{queiroz_using_2025}. Current methodologies attempting to address fairness without demographic metadata face fundamental limitations, including systematic biases, accuracy limitations, technical barriers, and compromised transparency, thus emphasizing the critical need for rigorous evaluation protocols and explicit methodological guidelines \cite{ashurst_fairness_2023}.

\textbf{Text and image.} Recent research on VLMs highlights that data curation plays a more critical role in CLIP model performance than the architecture itself \cite{xu_demystifying_2023}. While enhancements such as the addition of sigmoid loss contribute to performance \cite{zhai_sigmoid_2023a}, the success of VLMs primarily depends on abundant high-quality data and substantial compute resources for scaling \cite{tschannen_siglip_2025}. The process of curation leverages text metadata to construct more relevant datasets by selecting key terms, applying substring matching, and most importantly, balancing the dataset through sub-sampling based on chosen metadata \cite{xu_demystifying_2023}.

\textbf{Scalable methods without protected attributes.} FMs serve as highly effective feature extractors, enabling clustering-based approaches for automatic data curation that do not require protected attributes. This capability facilitates the enhancement of diversity and balance in large datasets \cite{vo_automatic_2024, queiroz_using_2025}. When compared to uncurated data, features derived from these automatically curated datasets demonstrate superior performance \cite{vo_automatic_2024}, for example by reducing a 4.44\% gender imbalance to achieve an equal distribution of male and female samples in medical imaging \cite{queiroz_using_2025}. The clustering techniques developed through this approach serve a dual purpose: they not only improve data quality but also provide a systematic method for identifying potential biases in models \cite{queiroz_using_2025}. These findings establish FMs as versatile tools that excel not only in feature extraction but also in crucial data management tasks, including bias assessment and automated curation.

\textbf{Data curation alone is insufficient.} While data balancing contributes to model fairness, it does not fully eliminate inherent biases in AI systems. The integration of multiple data domains, particularly the combination of textual and visual data, can potentially introduce or intensify biases beyond those present in unimodal models \cite{hall_visionlanguage_2023, hutchinson_underspecification_2022, booth_bias_2021}. Consequently, achieving fair downstream behavior requires comprehensive mitigation strategies beyond data balancing alone \cite{alabdulmohsin_clip_2024}. In the context of unimodal models, performance disparities in downstream tasks frequently correlate with class difficulty, where more challenging classes exhibit higher misclassification rates and decreased performance metrics \cite{cui_classes_2024}.

\section{Enviromental Impact}

This section examines the environmental considerations throughout FMs development, encompassing training requirements, model evaluation protocols, and deployment strategies.

\subsection{Training}
\label{sec:training}
\textbf{Fairness.} Self-supervised learning models leveraging diverse, unlabeled datasets demonstrate enhanced fairness and inclusivity compared to supervised approaches, yielding outcomes characterized by increased robustness and reduced bias \cite{goyal_vision_2022}. However, FMs retain inherent biases despite these advantages \cite{glocker_risk_2023, vaidya_demographic_2024, luo_fairclip_2024}. For example, in a chest radiograph classification task, female patients experienced a 6.8–7.8\% decrease in performance on the “no finding” label \cite{glocker_algorithmic_2023}. The mitigation of such biases necessitates targeted interventions within the training loop. This optimization process faces dual challenges: the substantial computational requirements and the inherent difficulty of acquiring large-scale datasets that exclude protected attributes.

\textbf{Computational Challenges.} The substantial computational requirements of FMs arise from two primary factors: the necessity of extensive datasets during pre-training and the complexity of architectures involving numerous parameters. These resource demands create significant entry barriers, constraining many institutions to fine-tuning existing pre-trained models rather than developing their own. Additionally, as established in the Data Creation section, this computational divide disproportionately affects certain countries, potentially exacerbating existing biases and societal inequities.

\textbf{Bias.} The lack of data representativeness, as illustrated in Figure \ref{fig:map}, can cause models to perform disproportionately better on more prevalent or overrepresented data subsets, thereby limiting generalizability \cite{cui_classes_2024}. Models are known to internalize diverse biases related to complexity, shape, and other intrinsic features present in training data, which compromises their robustness \cite{stanley_where_2025, lampinen_learned_2024}. Hence, assembling datasets that comprehensively reflect the true population distribution is paramount to enhancing model robustness and mitigating bias-related performance degradation during training.

\textbf{Bias Amplification in VLMs.} Multimodal models not only perpetuate existing societal biases but also amplify them relative to single-modality systems \cite{hutchinson_underspecification_2022, hall_visionlanguage_2023, booth_bias_2021}. The pairing of images and text reintroduces the well-documented problem of label bias \cite{wang_enhancing_2024, chen_what_2024}, and generating non-factual medical diagnoses and overconfidence in generated diagnoses \cite{NEURIPS2024_fde7f40f}. Moreover, as shown in the Table \ref{tab:medical-datasets}, most multimodal datasets remain English-centric, which restricts their applicability in global contexts \cite{bassignana_ai_2025}. This limitation is reinforced by the fact that most multilingual solutions rely heavily on translation pipelines, wich could introduce semantic drift and cultural misalignment \cite{derakhshani_neobabel_2025}, in the medical domain this could lead to a misdiagnosis of local disease.

\textbf{Synthetic data.} Integrating synthetically generated data during training offers a promising approach to enhancing fairness in FMs. Evidence demonstrates that models trained exclusively on synthetic data can match or exceed the accuracy of models trained with real data \cite{khosravi_synthetically_2024, moroianu_improving_2025, ktena_generative_2023}. For instance, in chest radiography applications, models trained with synthetic data achieved a 6.5\% accuracy increase in downstream classification performance compared to models trained with real data \cite{moroianu_improving_2025}.
Models that integrate medical findings with patient demographics prove essential for generating well-distributed synthetic datasets, as exemplified by RoentGen-v2 \cite{moroianu_improving_2025}.
Model performance depends critically on synthetic data quality. Traditional high-quality generation methods require substantial computational resources. Emerging techniques such as flow matching and consistency models address this limitation through end-to-end architectures that reduce computational demands while maintaining generation quality \cite{geng_mean_2025, song_improved_2023}.

\subsubsection{Pre-training}
\textbf{Protected attributes.} The computational challenges inherent in vision and language models stem from power-law scaling relationships, wherein incremental performance improvements necessitate exponential increases in computational resources \cite{goyal_selfsupervised_2021, goyal_vision_2022}. As established in the Data Creation section, the prevalent absence of metadata in large-scale datasets introduces significant complexities for training loop implementation in pre-trained FMs. These constraints fundamentally limit the application of fairness techniques during the training process. Furthermore, fairness methodologies developed for pre-trained FMs must demonstrate efficient scalability across expansive datasets and sophisticated architectural frameworks.

\textbf{Data selection.} The incorporation of systematic data curation into the training loop can be achieved through active selection strategies, wherein computational priority is assigned to data elements that maximize task performance contributions \cite{schaul_prioritized_2015}. Research demonstrates that small curated models can facilitate the training of larger models through systematic identification of both straightforward and challenging image cases \cite{evans_bad_2024, evans_data_2024}. Implementation of these selection methodologies yields dual benefits: reduced training duration for pre-trained models and enhanced quality of resultant outputs.
 
\textbf{Loss.} The Reducible Holdout Loss (RHO) employs a secondary model to identify three categories of data points: those that are learnable, those worth dedicating computational resources to learn, and those not yet acquired by the model \cite{mindermann_prioritized_2022}. One established approach for enhancing model robustness and fairness involves the integration of protected attributes into loss functions, thereby promoting equitable outcomes across demographic groups \cite{mandal_ensuring_2020}. However, as examined in the Data Documentation section, the acquisition of protected attributes presents significant practical challenges. To address this limitation, alternative methodologies utilizing clustering-based data curation \cite{vo_automatic_2024} offer potential proxy measures for protected attributes \cite{queiroz_using_2025}.

\textbf{World Models.} Contemporary research in robust and unbiased model development prioritizes the prediction of representations within embedding spaces over traditional token-level forecasting approaches (e.g., word or pixel prediction) \cite{lecun_path_}. Within this framework, a novel class of FMs, termed World Models enables simultaneous training across multiple languages and modalities while maintaining scalability and minimizing bias \cite{team_large_2024}. In the visual domain, the I-JEPA implements this embedding-centric approach, yielding substantial improvements in robustness, scalability, and computational efficiency relative to MAE \cite{assran_selfsupervised_2023, littwin_how_2024}.

\textbf{Vision Language World Models.} Current VLMs, such as LLaVA-Med \cite{li_llavamed_2023} and MedVInT \cite{zhang_pmcvqa_2023}, exhibit notable challenges including overconfidence in diagnostic outputs, privacy breaches, and the perpetuation of health disparities, as mention in Section \ref{sec:training} \cite{NEURIPS2024_fde7f40f}. A promising approach to mitigate these issues involves integrating language and vision within world models equipped with a planning mechanism \cite{chen_planning_2025}. This mechanism operates by evaluating the consequences of an action, generating multiple potential outcomes, and predicting the optimal course that minimizes a cost function as assessed by a critic model \cite{chen_planning_2025, lecun_path_}. Crucially, fairness constraints can be embedded into this critic model. As this component is a self-supervised language model, establishing robust fairness principles in the language domain becomes fundamental to mitigating systemic bias throughout the entire vision-language architecture.

\textbf{Knowledge Agglomeration.} Knowledge agglomeration has recently emerged as a promising approach for pre-training. This approach leverages knowledge from existing FMs with diverse perspectives, such as CLIP \cite{alabdulmohsin_clip_2024}, DINO \cite{oquab_dinov2_2023}, and SAM \cite{kirillov_segment_2023}, to train improved models. Recent examples, like RADIO \cite{heinrich_radiov25_2025}, demonstrate that models trained via knowledge agglomeration can surpass their teacher's performance. In the medical imaging domain, models such as MedSAM \cite{ma_segment_2024} and MedCLIP \cite{wang_medclip_2022} can be combined to enhance performance. Crucially, this approach allows integrating models with fairness-aware representations alongside those optimized for general performance, jointly improving subgroup representation while balancing the fairness-utility trade-off (see Section \ref{Model Evaluation} for a detailed discussion of utility-fairness trade-offs).

\subsubsection{Fine-tuning}
\label{sec:fine-tuning}
\textbf{Efficacy of Fine-tuning for Fairness Enhancement.} Fine-tuning serves as a critical mechanism for aligning pre-trained FMs with targeted objectives. These models demonstrate exceptional capability in adapting to novel data distributions with minimal sample requirements \cite{azizi_robust_2023}. Fine-tuning effectively mitigates bias through increased model sensitivity to training distribution, particularly when employing balanced and curated datasets \cite{alabdulmohsin_clip_2024}.

\textbf{Resource Optimization in Fine-tuning.} In contrast to the computationally intensive pre-training phase, fine-tuning procedures can be executed with substantially reduced resource requirements, rendering this approach particularly advantageous for resource-constrained initiatives. The utilization of domain-specific pre-trained FMs, especially those optimized for medical imaging applications, further enhances computational efficiency in settings with limited infrastructure.

\textbf{Parameter-efficient Fine-tuning.} Methodologies such as LoRa \cite{hu_lora_2021} and QLoRa \cite{dettmers_qlora_2023} enhance the efficiency of low-resource fine-tuning through selective parameter modification, targeting only specific subsets within the base model architecture. The impact of these optimization techniques on bias mitigation remains an active area of investigation, with current research providing inconclusive evidence regarding their effects on model fairness \cite{ding_fairness_2024, jin_fairmedfm_2024}. Notably, alternative computational optimization strategies, including model pruning and differentially private training approaches, demonstrate increased bias manifestation within specific demographic subgroups \cite{tran_pruning_2022, bagdasaryan_differential_2019}.

\textbf{Mitigating Bias.} Despite extensive research into bias mitigation strategies within deep learning frameworks, conventional fairness interventions demonstrate variable efficacy when applied to FMs \cite{jin_fairmedfm_2024}. The limitations of these approaches extend beyond FMs applications, with traditional deep learning implementations often achieving only modest improvements in fairness metrics \cite{zong_medfair_2023}. Advanced data augmentation strategies, including AutoAug, Mixup, and CutMix methodologies, demonstrate promise in addressing challenging FMs scenarios \cite{cui_classes_2024}. Furthermore, the integration of synthetically generated data during the fine-tuning process presents compelling evidence for enhanced fairness outcomes \cite{ktena_generative_2023}.

\textbf{Blackbox FMs.} The increasing commercialization of FMs and proliferation of their APIs has resulted in a distribution model where access is frequently restricted to embedding outputs, limiting direct interaction with model architectures. This constrained accessibility presents distinct challenges for bias and hallucinations mitigations in medical imaging applications. Recent research demonstrates, however, that effective bias and hallucinations eliminations techniques can be implemented without requiring access to internal model parameters, offering viable solutions for both open-source and proprietary model frameworks \cite{jin_universal_2024, manakul_selfcheckgpt_2023}.

\subsection{Model Evaluation}
\label{Model Evaluation}
\textbf{Fairness Considerations and Metrics.} Within deep learning research, fairness has become a critical consideration, with particular significance in medical imaging applications \cite{chen_algorithmic_2023, xu_addressing_2024, riccilara_addressing_2022, shi_survey_2024, vaidya_demographic_2024}. The evaluation of fairness encompasses two fundamental methodological approaches: individual fairness, which requires consistent model outputs for similar inputs, and group fairness, which assesses model performance across demographic categories defined by protected attributes such as race and gender. Despite the widespread adoption of group fairness metrics in practical applications, a significant number of influential FMs studies in medical imaging have conducted evaluations without incorporating explicit fairness metrics \cite{azizi_robust_2023, zhou_foundation_2023, lu_visuallanguage_2024, zhao_foundation_2024}, thus leaving critical questions about potential biases and their clinical implications unexplored.

\textbf{Utility-Fairness Trade-offs.}
In real-world scenarios, models may use protected attributes as shortcuts for specific tasks, introducing bias \cite{yang_limits_2024}. However, bias does not always stem from such shortcuts; other factors like intensity-based and morphology-based effects can also influence model behavior \cite{stanley_where_2025}. This bias complexity creates a utility-fairness trade-off, where achieving statistical parity can reduce utility for some groups \cite{zhao_inherent_2022, wei_fairnessaccuracy_2021}. Identifying the optimal trade-off point is essential to balance both metrics effectively \cite{dehdashtian_utilityfairness_2024, ozbulak_multiobjective_2025}.

\textbf{Hallucinations.} Medical hallucinations arise in specialized tasks such as diagnostic reasoning, employing domain-specific terminology that masks critical inaccuracies\cite{kim_medical_2025}. Expert scrutiny remains essential for detecting these plausible yet incorrect outputs, which can delay appropriate interventions and compromise patient safety \cite{kim_medical_2025}. Large language model taxonomies typically categorize hallucinations into factual errors, outdated references, spurious correlations, incomplete reasoning chains, and fabricated sources across text and multimodal contexts \cite{kim_medical_2025}. Detecting and quantifying hallucinations is critical for ensuring model safety and reliability \cite{chen_detecting_2024, kim_medical_2025, gu_medvh_2024}. Exploring the relationship between protected attributes and hallucinations represents an emerging research area that advances fairness evaluations and addresses bias in generative models \cite{moroianu_improving_2025, kim_medical_2025}. In this context, rigorous quality checks of generated images are essential to remove hallucinations and mitigate bias; for example, in a dataset of 623,712 prompts, 58,558 (9.4\%) failed quality control, predominantly due to violations of race-related criteria \cite{moroianu_improving_2025}.

\textbf{Robustness.} A fundamental challenge in medical imaging applications lies in ensuring that fairness-aware models maintain their equitable performance when transitioning between different data distributions (A to B), particularly given the dynamic nature of population characteristics and deployment contexts \cite{schrouff_diagnosing_2023}. Although adversarial training techniques enhance overall model robustness, the improvements typically manifest unevenly across different classes \cite{xu_be_2021, ma_tradeoff_2022}. Research indicates that fairness enhancement strategies can simultaneously strengthen model robustness \cite{mou_fairness_2024}. However, contemporary deep learning systems frequently demonstrate inadequate robustness to protected attributes, specifically sex and gender variables \cite{zong_medfair_2023}. Within FMs frameworks, the direct deployment of models without fine-tuning procedures further compromises robustness to these demographic attributes, thus constraining their efficacy across diverse application scenarios \cite{queiroz_using_2025}.

\textbf{Data-Efficient Generalization.} FMs demonstrate exceptional capabilities in data-efficient generalization, facilitating fine-tuning processes for medical imaging tasks with minimal labeled data requirements \cite{azizi_robust_2023}. This characteristic proves particularly advantageous for resource-constrained institutions implementing model deployment within their specific data environments. Nevertheless, substantial uncertainty persists regarding the implications of limited labeled data usage on fairness metrics in FMs applications. The development of ethical frameworks and bias mitigation strategies assumes critical importance in these contexts, especially given that institutions operating under resource constraints typically depend on restricted labeled data availability.

\textbf{Benchmark.} Benchmarks are essential for assessing fairness in FMs development, providing teams with systematic evaluation tools. While fairness benchmarks exist for deep learning in medical imaging \cite{zong_medfair_2023, zhang_improving_2022, zhou_radfusion_2021, dutt_fairtune_2024}, comprehensive benchmarks and libraries specifically designed for FMs remain lacking \cite{jin_fairmedfm_2024, khan_how_2023}. Developing such resources is crucial for holistic fairness evaluation, encompassing metrics, utility, utility-fairness trade-offs, robustness, and data-efficient generalization. Furthermore, FMs should undergo testing on unbiased data and pipelines, with evaluations conducted on distributions that reflect real-world applications to ensure valid and applicable fairness metrics \cite{longpre_responsible_2024}. Benchmarks such as Med-HallMark, which introduces detection methods including MediHallDetector, provide systematic frameworks for evaluating and mitigating hallucinations in medical applications \cite{chen_detecting_2024}.

The Fairness Benchmarking for Medical Imaging Foundation Models (FairMedFM) \cite{jin_fairmedfm_2024} provides a comprehensive framework for comparing models, techniques, and datasets, covering most of the aspects discussed here. Given the importance of fairness and the lack of protected attributes in many datasets (see Table~\ref{tab:medical-datasets}), we encourage both new and existing datasets to collect this information. Even if such metadata is limited to the test set or a reduced patient cohort, it should accurately reflect the dataset distribution to support fairness evaluation and model development, as demonstrated by datasets like Harvard-FairVLMed \cite{luo_fairclip_2024}. Legacy datasets such as PMC-15M \cite{zhang_biomedclip_2025}, which are significant for VLMs research, could be enhanced by extracting demographic metadata such as age and sex from captions. Methods for dataset curation like MetaCLIP \cite{xu_demystifying_2023} can facilitate this process, improving both evaluation and model training.  In cases where protected attributes cannot be obtained, we recommend techniques that create groups approximating demographic attributes to facilitate fairness analysis \cite{queiroz_using_2025,vo_automatic_2024}.

\subsection{Deployment}
\textbf{Documentation.} Comprehensive documentation constitutes a fundamental requirement for the ethical implementation of FMs \cite{longpre_responsible_2024, bommasani_foundation_2023}. The documentation must encompass detailed specifications of training data sources, testing benchmark methodologies, and explicit deployment guidelines that minimize risk and bias \cite{mitchell_model_2019}. Moreover, the documentation should present information in a stratified manner, ensuring accessibility across varying levels of technical expertise while maintaining rigorous detail. This multi-level documentation approach enables stakeholders to comprehend the model's inherent risks, biases, and operational constraints. Such transparency serves as a crucial mechanism for fostering trust and promoting responsible model deployment in clinical settings.

 \textbf{License.} Open FMs represent a strategic approach to addressing global bias mitigation, enhancing transparency, and facilitating equitable power distribution \cite{bommasani_considerations_2024, matheny_artificial_2025}. By providing unrestricted access to data, code, and model weights, this approach enables resource-constrained institutions to engage in model development and adaptation, thereby fostering innovation through reduced computational barriers. As established in Section \ref{sec:fine-tuning}, while closed models present substantial impediments to achieving fairness and impartial outcomes, open-weight architectures facilitate superior downstream task adaptation. Nevertheless, ensuring adherence to intended model applications remains crucial for preventing bias propagation and maintaining equity. The implementation of a Responsible AI License framework provides a structured mechanism for guiding ethical and equitable model utilization. 

\textbf{Monitoring.} In medical settings, the prevalence of the disease and the distribution of patients accessing a specific hospital can change over time. While assessing fairness and bias during pretraining and fine-tuning is critical for FMs, continuous monitoring of deployed models in real-world scenarios is equally important. Despite its importance, such monitoring is not widely practiced; in the United States, only 44\% of institutions reported conducting local evaluations for bias \cite{nong_current_2025}.
\section{Policymakers}

\begin{figure*}[h]
    \centering
    \includegraphics[width=1\textwidth]{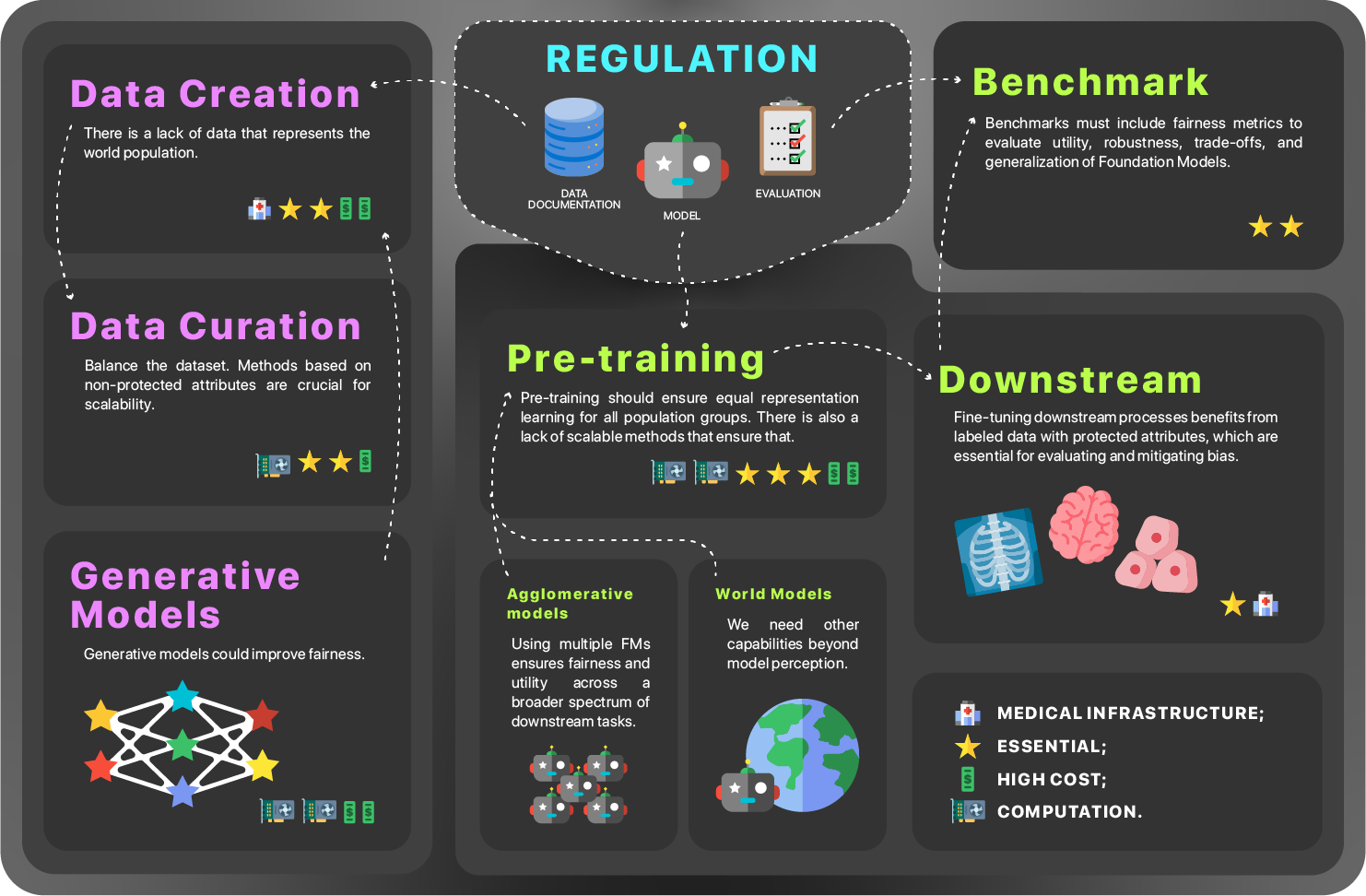}
    \caption{\textbf{Overview recommendation:} The figure summarizes recommendations for achieving fair foundation models, indicating for each stage its medical infrastructure, importance for fairness (essential), high cost, and computation. The recommendations focus on where governments and industry should prioritize technical interventions to advance fairness. Medical infrastructure is critical for both downstream applications and data creation, but their cost profiles differ: data creation depends primarily on expensive imaging equipment and large-scale data storage, whereas downstream tasks require substantial clinical expertise for data annotation. With respect to fairness mitigation, the most critical and challenging stage is pre-training, which demands extensive computational infrastructure and incurs high costs. The figure also highlights evaluation and data curation as comparatively low-cost yet important strategies for improving fairness. Generative models offer a promising avenue to balance data creation and alleviate infrastructure constraints by better representing the underlying population. At the same time, real-data collection remains more critical for fairness, because equitable access to medical infrastructure across all population groups is essential, whereas generative models can only complement these efforts by improving population representation in the training data. Finally, agglomerative models and world models are identified as future directions with potential to enhance fairness, although they are tightly coupled to pre-training and therefore are not quantitatively assessed in the figure.}
    \label{fig:summary_recommendation}
\end{figure*}

    This section examines two critical aspects of FMs development: governance frameworks for ensuring reliability and the strategic allocation of essential resources. Figure~\ref{fig:summary_recommendation} summarizes the key points discussed in the text, providing a comprehensive overview of foundation model development along with the cost and importance of each step.

\subsection{Governance}
\textbf{Reliability.} The implementation of FMs in healthcare faces critical reliability challenges due to insufficient standardization in bias and hallucinations detection and mitigation practices \cite{khan_how_2023, glocker_risk_2023, li_empirical_2024, jin_fairmedfm_2024, zong_medfair_2023}. Despite a growing emphasis on AI ethics, healthcare institutions lack both established evaluation frameworks and qualified experts needed to assess these systems effectively. This deficiency is particularly concerning because bias continues to emerge in deployment scenarios, while current mitigation methods demonstrate limited reliability \cite{jin_fairmedfm_2024, zong_medfair_2023}. The challenge of bias mitigation becomes particularly critical in regions with limited data representation, notably in Global South nations (Figure \ref{fig:map}), where healthcare systems encounter substantial obstacles in implementing and adapting these models for their populations. This disparity underscores a critical gap in current FMs deployment strategies, as the absence of representative data further compounds existing healthcare inequities.

\textbf{FMs as institutions.}
Research demonstrates that algorithmic systems function as institutional frameworks that organize complex machine-human interactions in decision-making processes \cite{almeida_algorithms_2022}. FMs represent a significant evolution in this infrastructure, enabling complex decision-making capabilities across medical domains, offering pre-trained architectures adaptable to diverse medical tasks. However, these models systematically perpetuate societal power imbalances and biases through their operational frameworks \cite{cottier_rising_2024}. Conceptualizing FMs as institutions, rather than purely technical implementations, facilitates the establishment of comprehensive evaluation protocols and governance frameworks that enhance model reliability through systematic oversight, standardized assessment criteria, and robust accountability measures.

\textbf{Regulation.} Legislative frameworks are crucial in enhancing AI system reliability in healthcare. The EU AI Act exemplifies a pioneering approach by classifying medical AI systems as "high-risk" and establishing provisions for General Purpose AI Models (GPAI), which encompass FMs. These models present distinct challenges due to their scale, adaptability, and cross-domain impact \cite{minssen_challenges_2023}.
The EU AI Act mandates requirements to detect, prevent, and mitigate bias in applications that could result in discrimination. These regulatory requirements transform FMs development pipelines by necessitating the adoption of fairness-enhancing techniques discussed in this work. Although these changes impose extensive technical documentation requirements and associated costs \cite{aboy_navigating_2024, carl_impact_2024}, the adoption of such pipelines enhances the reliability of companie’s and government products and enables the scalable deployment of AI from local settings to national and global levels.

\subsection{Resource Allocation}

\textbf{Workforce.} Fair FMs demand interdisciplinary collaboration among technical, legal, and social experts to ensure comprehensive bias mitigation \cite{lekadir_futureai_2025, bengio_international_}. These multidisciplinary teams integrate diverse societal values and ethical perspectives into model development and evaluation protocols. Essential to this process is the active participation of marginalized communities, whose insights prove crucial for identifying systemic biases in FMs. The global distribution of FM expertise, however, remains heavily concentrated in the United States and China \cite{alshebli_china_2024, zwetsloot_skilled_2021}, creating significant barriers to diverse team formation. Another challenge arises from data sources and computing resources; many countries face difficulties in providing high-performance computing (HPC) access to institutions. A valuable contribution in this regard is the establishment of training centers that offer education, data, and computing resources.

\textbf{Data source.} AI research disparities manifest critically in data infrastructure and computational resources across global regions. Figure \ref{fig:map} illustrates the significant underrepresentation of Global South populations in major medical imaging datasets. Quantitative analyses reveal systematic exclusion across modalities, with Global South regions constituting less than 0.7\% of text-domain datasets \cite{longpre_bridging_2024}. This limitation extends to linguistic diversity: among medical imaging repositories, only PadChest provides non-English (Spanish) annotations (Table \ref{tab:medical-datasets}). Developing equitable FMs requires datasets that reflect global population diversity across both geographic and demographic dimensions. Achieving representative data collection necessitates integrated technical and governance frameworks that actively incorporate marginalized populations into healthcare data systems \cite{worldhealthorganization_conceptual_2010, chen_ethical_2021, bailey_structural_2017, williams_understanding_2019}.

\textbf{Computing.} FMs development requires extensive computational infrastructure, limiting access to organizations with advanced resources \cite{cottier_rising_2024}. Analysis of the November 2024 Top500 supercomputer rankings demonstrates pronounced regional disparities: North America holds 55.6\% of capacity, Europe 27.4\%, and Asia 15.9\%, while South America and Oceania represent just 0.6\% and 0.5\% respectively. The combined scarcity of computational power and large-scale datasets creates a fundamental barrier in the Global South, where only Brazil, Australia, and Argentina maintain Top500-listed facilities. Africa's absence from these rankings underscores how infrastructure limitations constrain both FMs development and deployment.

\section{Conclusion}

FMs represent a transformative advancement in medical imaging analysis, yet their implementation presents both opportunities and challenges for achieving equitable healthcare delivery. Our comprehensive review demonstrates that effective bias mitigation in FMs requires systematic interventions throughout the development pipeline, from data curation to deployment protocols. While technical innovations in training methodologies show promise for enhancing fairness without relying on protected attributes, the substantial computational and data demands of these models risk exacerbating global inequalities. The emergence of regulatory frameworks such as the EU AI Act reflects growing recognition of FMs societal impact and the need for governance structures that ensure responsible development. However, significant disparities persist, particularly in Global South nations where limited access to essential resources including specialized workforce, datasets, and computational infrastructure hinders both development and implementation of fair FMs. Moving forward, addressing these challenges requires coordinated action between technologists, healthcare providers, and policymakers to develop accessible solutions and appropriate frameworks for low-resource countries and institutions. As FMs continue to evolve, their successful implementation in healthcare will depend on our ability to balance technical innovation with ethical principles, ultimately working toward reducing rather than amplifying existing healthcare disparities.

\section*{Acknowledgements}
We thank Fundação de Amparo à Pesquisa do Estado de São Paulo (FAPESP), grants 21/14725-3 and 23/12493-3,
Conselho Nacional de Desenvolvimento Científico e Tecnológico (CNPQ), Swiss National Science Foundation
(SNSF) under Grant No. 200021E\_214653, Santos Dumont supercomputer at the LNCC. We would like thank Lucas Tosta for the designs.

\bibliographystyle{unsrt}
\bibliography{global-fairness-fms.bib}  






\end{document}